\def\year{2021}\relax
\documentclass[letterpaper]{article} 
\usepackage{aaai21}  
\usepackage{times}  
\usepackage{helvet} 
\usepackage{courier}  
\usepackage[hyphens]{url}  
\usepackage{graphicx} 
\usepackage{natbib}  
\usepackage{caption} 
\usepackage{booktabs}       
\usepackage{amsmath}
\usepackage{amssymb}

\title{Residual Shuffle-Exchange Network \\ for Fast Processing of Long Sequences}
\author {
    Andis Draguns, Em\={\i}ls Ozoli\c{n}\v{s}, Agris \v{S}ostaks, Mat\={\i}ss Apinis, K\={a}rlis Freivalds \\
}

\affiliations{
    $\{$andis.draguns, emils.ozolins, agris.sostaks, matiss.apinis, karlis.freivalds$\}$@lumii.lv\\
    Institute of Mathematics and Computer Science,
    University of Latvia \\
}

\begin{document}
\maketitle


\begin{abstract}
Attention is a commonly used mechanism in sequence processing, but it is of O($n^2$) complexity which prevents its application to long sequences. The recently introduced neural Shuffle-Exchange network offers a computation-efficient alternative, enabling the modelling of long-range dependencies in O($n$ log $n$) time. The model, however, is quite complex, involving a sophisticated gating mechanism derived from the Gated Recurrent Unit. In this paper, we present a simple and lightweight variant of the Shuffle-Exchange network, which is based on a residual network employing GELU and Layer Normalization. The proposed architecture not only scales to longer sequences but also converges faster and provides better accuracy. It surpasses the Shuffle-Exchange network on the LAMBADA language modelling task and achieves state-of-the-art performance on the MusicNet dataset for music transcription while being efficient in the number of parameters. We show how to combine the improved Shuffle-Exchange network with convolutional layers, establishing it as a useful building block in long sequence processing applications.
\end{abstract}


\section{Introduction}
\label{introduction}
More and more applications of sequence processing performed by neural networks require dealing with long inputs. A key requirement is to allow modelling of dependencies between distant parts of the sequences. Such long-range dependencies occur in natural language when the meaning of some word depends on other words in the same or previous sentence. There are important cases, e.g., to resolve coreferences, when such distant information may not be disregarded.

In music, dependencies occur on several scales. At the finest scale samples of the waveform correlate to form note pitches, at medium scale neighbouring notes relate to each other by forming melodies and chord progressions, at coarse scale common melodies reappear throughout the entire piece creating a coherent musical form \citep{thickstun2016learning, huang2018music}. Dealing with such dependencies requires processing very long sequences (several pages of text or the entire musical composition) in a manner that aggregates information from their distant parts. Especially challenging are approaches that work directly on the raw audio waveform.

The ability to combine distant information is even more important for algorithmic tasks where each output symbol typically depends on every input symbol. The goal of algorithm induction is to derive an algorithm from input-output examples which are often given as sequences. Algorithmic tasks are especially challenging due to the need for processing sequences of unlimited length. Also, generalization plays an important role since training is often performed on short sequences but testing on long ones.

The commonly used attention mechanism (for example, in Transformers) can deal with such long-range dependencies but its time and space complexity is quadratic depending on the sequence length, therefore, is not an attractive choice for long sequences. The attention mechanism's complexity also makes it slower at inference time, making it less suitable for tasks with strict latency requirements, such as realtime sound processing. The recently introduced neural Shuffle-Exchange networks allow modelling of long-range dependencies in sequences in O($n$ log $n$) time \cite{shuffle-exchange}. The idea is very promising and offers a computation-efficient alternative to the attention mechanism. However, the model is quite complex, involving a sophisticated gating mechanism derived from the Gated Recurrent Unit \citep{cho2014learning}.

In this paper, we present a much simpler and faster version of the neural Shuffle-Exchange network which is based on the residual network idea employing Gaussian Error Linear Units \citep{hendrycks2016gaussian} and Layer Normalization \citep{ba2016layer} instead of gates.

We empirically validate our improved model on algorithmic tasks, LAMBADA question answering and multi-instrument musical note recognition (MusicNet dataset). It surpasses the original Shuffle-Exchange network by 2.1\% on the LAMBADA language modelling task and achieves state-of-the-art 78.02\% average precision score on MusicNet.

We introduce a modification where we prepend our proposed model with strided convolutions to increase the speed and applicability to long sequences even more. This change enables processing a sequence of length 2M symbols in only 3.97 seconds.

Our main contributions are:
\begin{itemize}
  \item We propose a much simpler and faster Switch Unit -- the core part of the Shuffle-Exchange network. This improvement leads to higher accuracy and scaling to longer sequences. It also makes the model approximately 4 times faster to train and 2 times faster on inference.
  \item We surpass the previous state-of-the-art on MusicNet while being efficient in the number of parameters. Our proposed improvements to the architecture enable this state-of-the-art achieving model to run inference on a single GPU fast enough to be suitable for realtime audio processing.
\end{itemize}

\section{Related Work}

The attention mechanism \citep{bahdanau2014neural} has become a standard choice in numerous neural models, including Transformer \citep{vaswani2017attention} and BERT \citep{devlin2018bert} which achieved state-of-the-art accuracy in NLP and related tasks. However, the complexity of the attention mechanism is quadratic depending on the input length and does not scale to long sequences.

One way to overcome the complexity of attention is cutting the sequence into short segments and using attention only within the segment boundaries \citep{al2018character}. Various sparse attention mechanisms have been proposed to deal with the quadratic complexity of dense attention by attending only to a small predetermined subset of locations \citep{beltagy2020longformer, zaheer2020big}. Reformer \citep{kitaev2020reformer} uses locality-sensitive hashing to approximate attention in time O($n$ log $n$). Linformer \citep{wang2020linformer} uses a linear complexity approximation to the original attention by creating a low-rank factorization of the attention matrix.

Another option for processing long sequences is using convolutional architectures \citep{gehring2017convolutional}. However, convolutions are inherently local $-$ the value of a particular neuron depends on a small neighbourhood of the previous layer. One way to capture long-range structure is to increase the receptive field of convolution by using dilated (atrous) convolution, where the convolution mask is spread out at regular spatial intervals. Dilated architectures have achieved great success in image segmentation \citep{yu2015multi} and audio generation \citep{oord2016wavenet}.

An important use of sequence processing models is in learning algorithmic tasks (see \citet{kant2018recent} for a good overview) where the way how memory is accessed is crucial. Neural GPU \citep{Kaiser2015NeuralGPU} utilizes active memory \citep{kaiser2016can} where computation is coupled with memory access. \citet{freivalds2017improving} proposes DNGPU where the flow of information in the Neural GPU is facilitated by introducing diagonal gates that improves training and generalization but does not address the performance problem caused by many layers.

For music transcription tasks convolutional architectures are common. See \citep{Benetos2019AutomaticMT} for a good overview. \citet{trabelsi2018deep} achieves notable performance on MusicNet by using a convolutional network based on complex numbers. \citet{yang2019complex} recently proposed a Transformer network that employs the Fourier transform of the waveform in the complex domain.

\section{Neural Shuffle-Exchange networks}

\begin{figure}[!b]
    \centering
    \includegraphics[width=0.7\columnwidth]{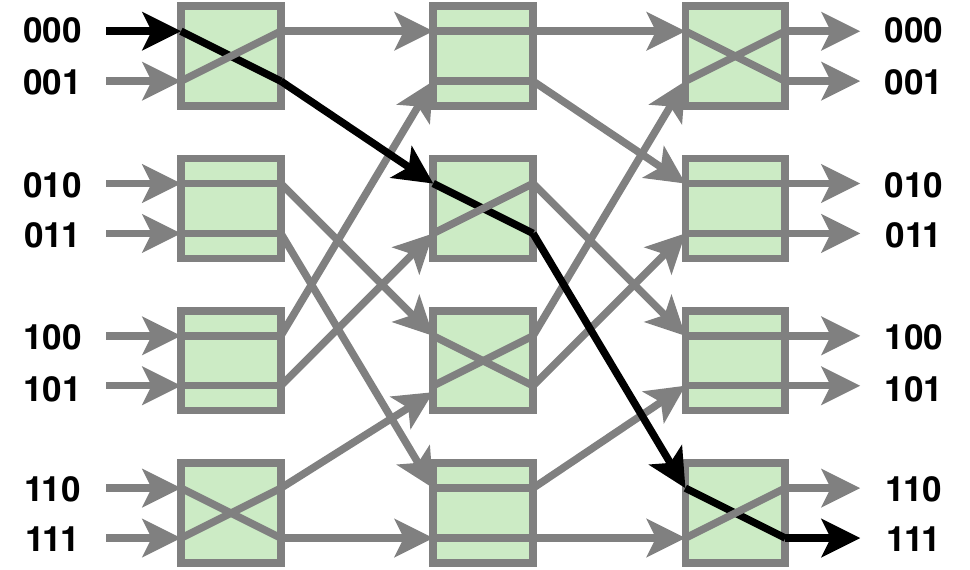}
    \caption{Shuffle-Exchange network routing signals from 8 input sequence addresses (sequence element locations) from the left to 8 output sequence addresses to the right. Each green block is a Switch Unit that takes two input elements and either swaps or leaves them unchanged. A column of Switch Units forms a Switch Layer. The arrows between two Switch Layers represent a Shuffle Layer that permutes the elements. In this figure, the switches are configured to route an element from address 000 to 111.
}
    \label{fig:shuffle-network}
\end{figure}

\begin{figure*}[!htp]
    \centering
    \includegraphics[width=2\columnwidth]{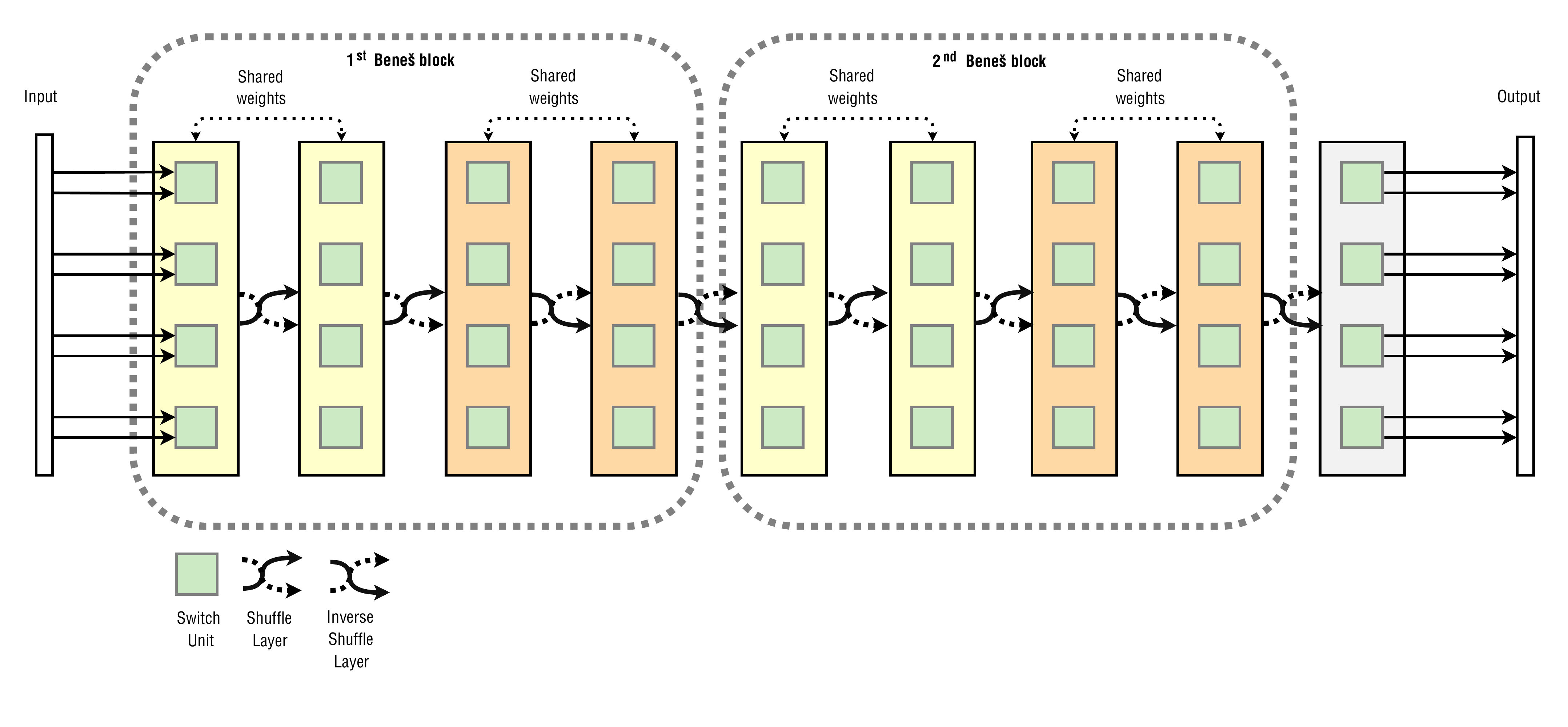}
    \caption{Residual Shuffle-Exchange network with two Bene\v{s} blocks and eight inputs. All learnable parameters are within the Switch Units. The rest of the network is fixed and used for routing information. Any number of Bene\v{s} blocks can be added to increase the depth of the model.
}
    \label{fig:sharing}
\end{figure*}

Neural Shuffle-Exchange network \cite{shuffle-exchange} has been recently proposed as an efficient alternative to the attention mechanism that allows modelling of long-range dependencies in sequences in O($n$ log $n$) time. The neural Shuffle-Exchange network is the neural adaption of the Shuffle-Exchange and the Bene\v{s} networks which are well-known from packet routing tasks in computer networks. These networks consist of interleaved shuffle and switch layers. The shuffle layer permutes the signals.
The switch layer consists of switches. Each switch is configured to either swap two adjacent signals or leave them unchanged. The neural Shuffle-Exchange network adapts the structure of Bene\v{s} network and replaces each switch with a Switch Unit -- a learnable 2-to-2 function. See Fig.~\ref{fig:shuffle-network} for an example of a neural Shuffle-Exchange network routing signals from 8 input sequence addresses to 8 output sequence addresses.

The input to the neural Shuffle-Exchange network is a sequence of length $n=2^k, k \in \mathbb{N}$, and each element of the sequence is a vector with $m$ dimensions. The input sequence is padded to the nearest power of two. The first layer of the model is a Switch Layer, which consists of $n/2$ Switch Units (SU). The input sequence is split into non-overlapping pairs of adjacent elements, and each pair is processed by a Switch Unit. Each Switch Unit is a neural network that computes a 2-to-2 function from a pair of elements. The second layer of the network is a Shuffle Layer, which permutes the inputs according to the perfect shuffle permutation. Perfect shuffle permutation maps each input address to an address that is circularly bit-shifted to the left (e.g. 101 to 011).

Note that all Shuffle Layers are identical and have no learnable parameters. It is also worth noting that both types of layers leave the dimensions of the input unchanged. The third layer of the network is the same as the first layer -- a Switch Layer. The layers keep alternating between Shuffle Layers and Switch Layers until there are a total of $\log(n)-1$ from each type of layer. This arrangement of layers constitutes the Shuffle-Exchange network.

The Shuffle-Exchange network is followed by a reversed Shuffle-Exchange network. The only difference between these two is that in the reversed Shuffle-Exchange network, all Shuffle Layers are replaced by inverse Shuffle Layers. Inverse Shuffle Layer permutes inputs like the regular Shuffle Layer, but the bit shift direction is to the right (e.g. 011 to 101). This combination of regular Shuffle-Exchange network followed by a reversed Shuffle-Exchange network forms a Bene\v{s} block -- a building block of the neural Shuffle-Exchange network. It has been shown \citep{dally2004principles} that such a Bene\v{s} block can connect any input to any output for each input-output simultaneously. Therefore, the neural Shuffle-Exchange network has a `receptive field' of the size of the whole sequence, and it has no bottleneck. These properties hold for dense attention but have not been shown for many sparse attention and dilated convolutional architectures. 

Multiple Bene\v{s} blocks can be stacked one after another to increase the depth of the model. After the last Bene\v{s} block, a final Switch Layer is added to complete the model. Within each Bene\v{s} block the weights are shared for each Switch Unit in the Shuffle-Exchange network and each Switch Unit in the reversed Shuffle-Exchange network (see Fig.~\ref{fig:sharing}). Such weight sharing enables generalization on algorithmic tasks because otherwise there would be no straightforward way of scaling up the model for sequences that are longer than the ones observed during training. No significant decrease in accuracy is observed on other tasks as a result of this weight sharing scheme.

At the heart of the neural Shuffle-Exchange network is the Switch Unit. The definition of the original Switch Unit from the neural Shuffle-Exchange network is:

\centerline{$\begin{aligned}
  s &= [s^1, s^2]\\
  r^1 &= \sigma(W_r^1s + B_r^1)\\
  r^2 &= \sigma(W_r^2s + B_r^2)\\
  c^1 &= \text{tanh}(W_c1(r^1 \odot s) + B_c^1)\\
  c^2 &= \text{tanh}(W_c2(r^2 \odot s) + B_c^2)\\
  u &= \sigma(W_us + B_u)\\
  \tilde{s} &= \text{swapHalf}(s^1, s^2)\\
  [s_o^1, s_o^2] &= u \odot \tilde{s} + (1 - u) \odot [c^1, c^2]
\end{aligned}$}

For a more detailed description of how this Switch Unit works, see \citet{shuffle-exchange}. This formulation of the Switch Unit uses sophisticated gating mechanisms similar to Gated Recurrent Unit \citep{cho2014learning} and is quite complex. Because apart from Switch Units, the network consists of only fixed non-trainable permutations, the choice of Switch Unit is critical to the overall performance of the network. Some alternatives to the Switch Unit have been explored, but so far the simpler architectures have led to a decrease in performance \citep{shuffle-exchange}.

\section{Residual Shuffle-Exchange Network}
\label{sec:model}

We propose the Residual Shuffle-Exchange network (RSE), which keeps the structure from the neural Shuffle-Exchange network but replaces the Switch Unit with our Residual Switch Unit (RSU). RSU is based on a residual network and employs Gaussian Error Linear Unit (GELU) \cite{hendrycks2016gaussian} and Layer Normalization. The unit's design is similar to the feed-forward block in the Transformer.

RSU takes as an input two vectors $[i_1, i_2]$ and produces two vectors $[o_1, o_2]$ as an output. Each of these vectors is of size $m$, where $m$ is the number of feature maps.

RSU consists of two linear transformations on the feature dimension. The first linear transformation is followed by Layer Normalization (LayerNorm) without output bias and gain \cite{xu2019understanding} and then by GELU. By default, we use a 2x larger hidden layer size than the input of the first layer, which is a good compromise of speed and accuracy (see Section~\ref{sec:ablation}). A second linear transformation is applied after GELU.
The RSU is defined as follows:

\centerline{$\begin{aligned}
i &= [i_1, i_2] \\
g &= \text{GELU}(\text{LayerNorm}(Z i))\\
c &= Wg + B\\
[o_1, o_2] &= \sigma(S)\odot i + h \odot c \\
\end{aligned}$}

In the above equations, $Z$, $W$ are weight matrices of size $2m \times 4m$ and $4m \times 2m$, respectively, $S$ is vector of size $2m$ and $B$ is a bias vector $-$ all of those are learnable parameters; $h$ is scalar value, $\odot$ denotes element-wise vector multiplication and $\sigma$ is the sigmoid function. For a visualization of the Residual Switch Unit, see Appendix \ref{sec:appendix-RSU}.

In this unit, we use a residual connection that is scaled by the learnable parameter $S$, which is restricted to the range [0,1] by the sigmoid function. Additionally, we scale the new value $c$ coming out of the last linear transformation by a constant $h$. We initialize $S$ and $h$ such that the signal travelling through the network keeps its expected amplitude at 0.25 under the assumption of the normal distribution (which is observed in practice). To have that, we initialize $S$ as $\sigma^{-1}(r)$ and $h$ as $\sqrt{1-r^2}*0.25$ where $r$ is an experimentally chosen constant close to 1. As $r$ approaches 1, RSU and RSE both start to behave more like identity functions. We use $r=0.9$, which works well. The signal amplitude after LayerNorm is 1, the weight matrix $W$ is initialized to keep this amplitude. If the amplitude of the input is 0.25, then the expected amplitude at the output is also 0.25, which is a good range for the softmax loss. During training, the network is free to adjust these amplitudes, but this initialization provides stable convergence even for deep networks.

Besides being simpler, the improved design of the RSU allows not using skip connections between Bene\v{s} blocks that were needed in the original SE network to ensure stable convergence. 

\subsection{Prepending convolutions}
There are tasks, e.g. the MusicNet task, where there is a large mismatch of information content between input and the Residual Shuffle-Exchange network -- each input unit contains one sample, but Residual Shuffle-Exchange network requires a large number of feature maps to work well. Encoding just one number into many feature maps is wasteful. For such tasks, we prepend the Residual Shuffle-Exchange network with several strided convolutions to increase the number of feature maps and reduce the sequence length. We use convolutions with stride 2 and apply LayerNorm and GELU after each convolution like in the RSU. Before the result is passed to the Residual Shuffle-Exchange network, a linear transformation is applied. An illustration of a concrete example model with two prepended convolution layers can be found in Appendix \ref{sec:appendix-musicnet-convolutions}.

Prepending convolutions shortens the input to the RSE network and speeds up processing. The obtained accuracy for the MusicNet is roughly the same. Note that this approach leads to a shorter output than the input. It may be necessary to append transposed convolution layers at the end of the network to upsample the signal back to its original length. For the MusicNet task, upsampling is not necessary since we utilize only a few elements of the output sequence.

\begin{figure*}[!ht]
    \centering
    \begin{minipage}[t]{.49\textwidth}
    \includegraphics[width=\columnwidth]{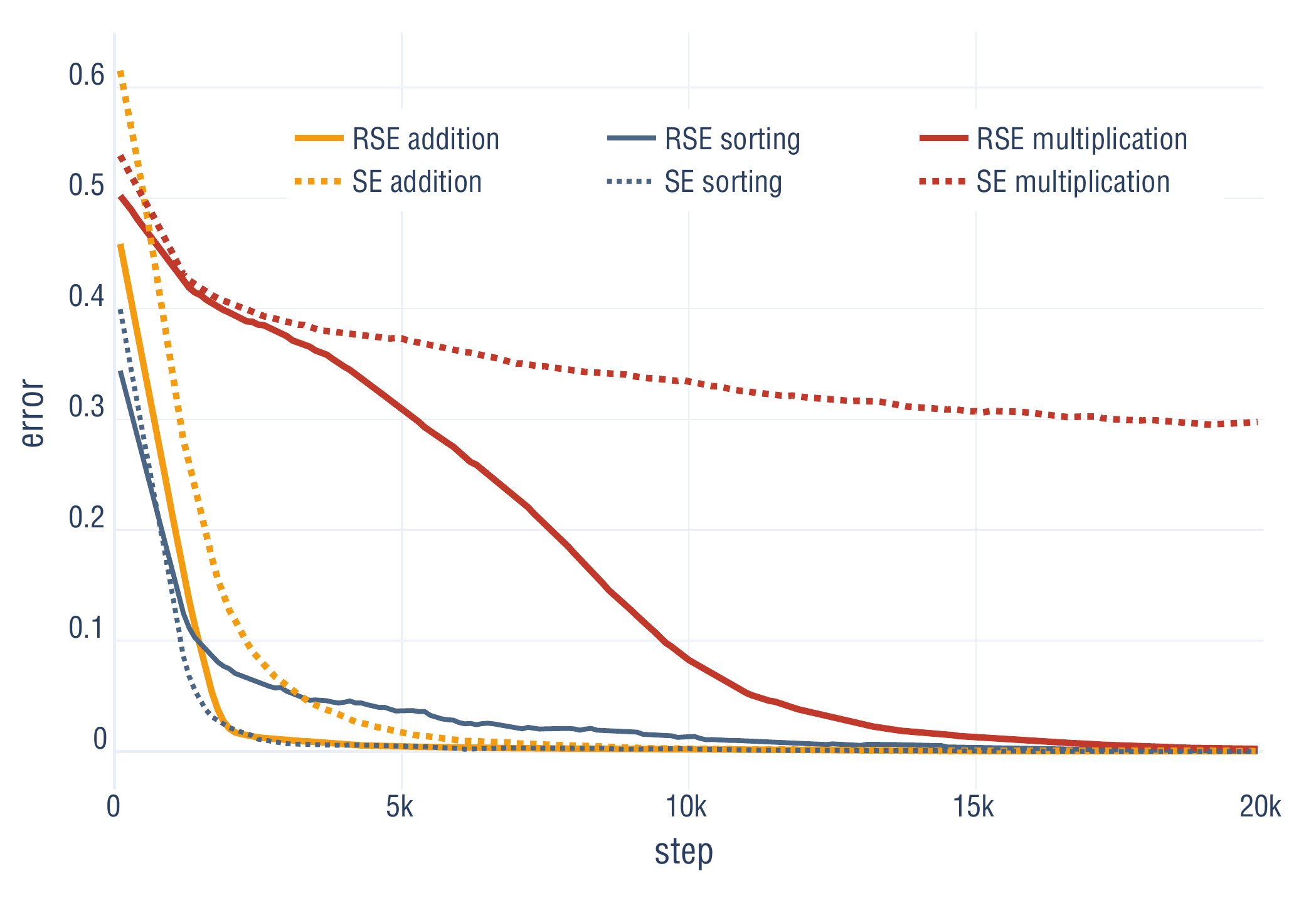}\hfill
    \caption{Test error depending on the training step for Residual Shuffle-Exchange (RSE) and Shuffle-Exchange (SE) networks on algorithmic tasks of length 64.}
    \label{fig:old-new}
    \end{minipage}%
    \hspace{.015\textwidth}
    \begin{minipage}[t]{.49\textwidth}
    \includegraphics[width=\columnwidth]{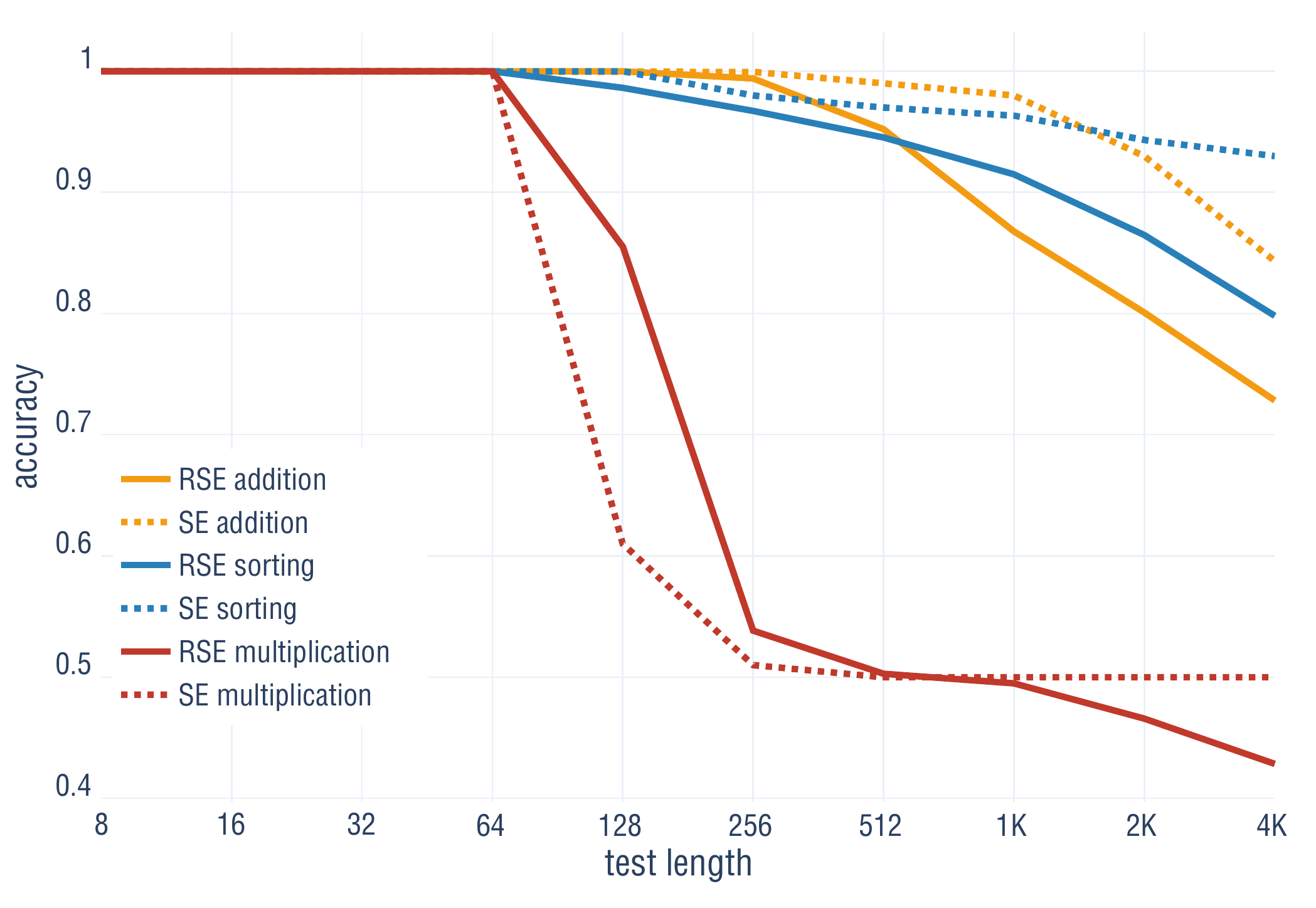}
    \caption{Test accuracy depending on the length for the generalization of Residual Shuffle-Exchange (RSE) and Shuffle-Exchange (SE) to longer sequences on algorithmic tasks.}
    \label{fig:generalization}
    \end{minipage}%
\end{figure*}


\section{Evaluation}
We have implemented the proposed architecture in TensorFlow. The code is at \url{https://github.com/LUMII-Syslab/RSE}. All models are trained on a single Nvidia RTX 2080 Ti (11GB) GPU with RAdam optimizer \citep{liu2019variance}

Our models were hand-tuned based on a coarse grid search of the parameters. Further increasing the model size leads to overfitting or negligible accuracy increase. For the other models, we use the hyperparameters that their authors have reported achieving the highest accuracy.

\subsection{Algorithmic tasks}

Let us evaluate how well the Residual Shuffle-Exchange (RSE) network performs on algorithmic tasks in comparison with the neural Shuffle-Exchange (SE) \citep{shuffle-exchange}. The goal is to infer O($n \log n$) time algorithms purely from input-output examples. In these tasks, a single bit change in the input can lead to a completely different output. Algorithmic tasks are good benchmarks to evaluate the model's ability to develop a rich set of long-term dependencies.

We consider long binary addition, long binary multiplication and sorting, which are common benchmark tasks in several papers including \citep{shuffle-exchange, kalchbrenner2015grid, zaremba2015reinforcement,zaremba2016learning,joulin2015inferring, grefenstette2015learning, Kaiser2015NeuralGPU, freivalds2017improving, dehghani2018universal}. 

The model for evaluation consists of an embedding layer where each symbol of the input is mapped to a vector of length $m$, one or two Bene\v{s} blocks and the output layer which performs a linear transformation to the required number of classes with a softmax cross-entropy loss for each symbol independently. We use the RSE model having one Bene\v{s} block for addition and sorting tasks, two blocks for the multiplication task and $m=192$ feature maps. 

We use dataset generators and curriculum learning from the article introducing neural Shuffle-Exchange networks \citep{shuffle-exchange}. For training, we instantiate several models for sequence lengths (powers of 2) from 8 to 64 sharing the same weights and train each example on the smallest instance it fits. We pad the sequence up to the required length with zeroes.
Figure \ref{fig:old-new} shows the testing accuracy on sequences of length 64 vs training step. We can see that on the multiplication task the proposed model trains much faster than SE, reaching near-zero error in about 20K steps vs 200K steps for the SE. For addition and sorting tasks, both models perform similarly.

We have compared the generalization performance of both models, see Fig.~\ref{fig:generalization}. We train both models on length up to 64 and evaluate on length up to 4K. On addition and sorting tasks, the proposed RSE model generalizes very well to length 256 but loses slightly to SE on longer sequences. For the multiplication task RSE model generalizes reasonably well to twice as long sequences but not more, where the old model does not generalize even this much. We have compared our model to the DNGPU -- the improved Neural GPU \citep{freivalds2017improving}. Our model outperforms it on all tasks except multiplication. More detailed comparison of DNGPU and RSE generalization performance can be found in Appendix \ref{sec:appendix-algorithms}.

\subsection{LAMBADA question answering}

The goal of the LAMBADA task is to predict a given target word from its broad context (on average, 4.6 sentences collected from novels). The sentences in the LAMBADA dataset \citep{Paperno2016TheLD} are specially selected such that giving the right answer requires examining the whole passage. In 81\% cases of the test set the target word can be found in the text, and we follow a common strategy \citep{Chu_2017, dehghani2018universal} to choose the target word as one from the text. The answer will be wrong in the remaining cases, so the achieved accuracy will not exceed 81\%. Choosing a random word from the passage gives 1.6\% test accuracy \citep{Paperno2016TheLD}.

We instantiate the model for input length 256 (all test and train examples fit into this length) and pad the input sequence to that length by placing the sequence at a random position and adding zeros on both ends. Randomized padding improves test accuracy. We use a pretrained fastText 1M English word embedding \citep{mikolov2018advances} for the input words. The embedding layer is followed by 2 Bene\v{s} blocks with 384 feature maps. To perform the answer selection as a word from the text, each symbol of the output is linearly mapped to a single scalar and we use softmax loss over the obtained sequence to select the position of the answer word.

In Table~\ref{lambada-comparison}, we show the test accuracy and the number of learnable parameters of our model in the context of results reported in previous works. The Residual Shuffle-Exchange network scores better than SE by 2.1\% while using 3x less learnable parameters \citep{shuffle-exchange}. Current state-of-the-art model GPT-3 surpasses our model, achieving 86.4\% accuracy while using about 16000 times more learnable parameters and pretraining on a huge dataset \cite{brown2020language}. The performance of GPT-3 is comparable to human performance on this task \citep{Chu_2017}. Our model scores lower than Universal Transformer by 1.66\% \citep{dehghani2018universal} but uses about 14 times fewer parameters. Our model also scores by 5.34\% higher than the Gated-Attention reader \citep{Chu_2017}.

\begin{figure*}[!ht]
    \centering
    \begin{minipage}[t]{.49\textwidth}
    \includegraphics[width=\columnwidth]{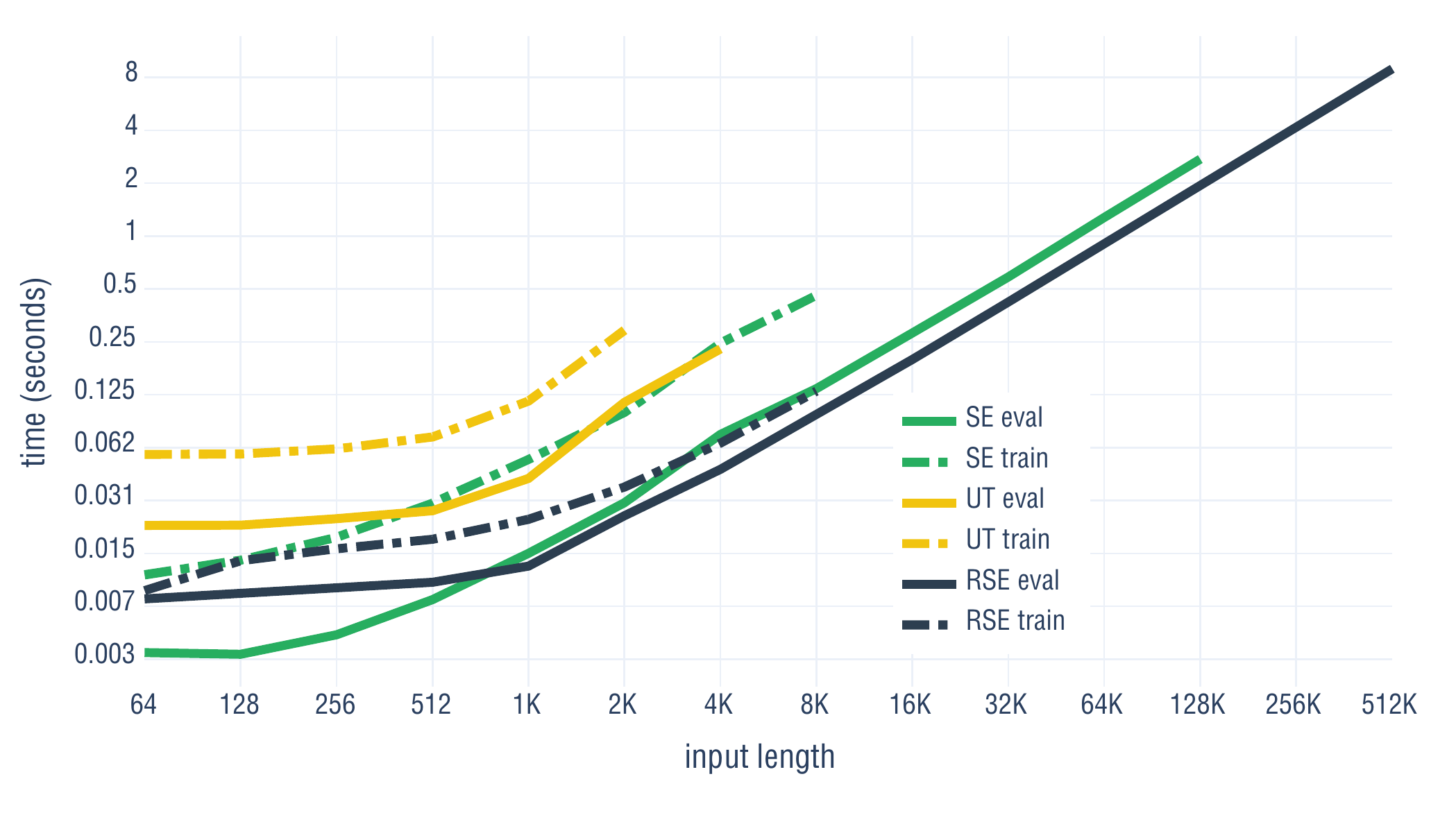}
    \caption{Evaluation and training time on different input lengths (log-scale). Each model is evaluated on sequence lengths that fit in the 11GB of GPU memory. RSE has about 4x faster training and 2x faster inference than SE.}
    \label{fig:lambada_time_comparison}
    \end{minipage}%
    \hspace{.015\textwidth}
    \begin{minipage}[t]{.49\textwidth}
    \includegraphics[width=\columnwidth]{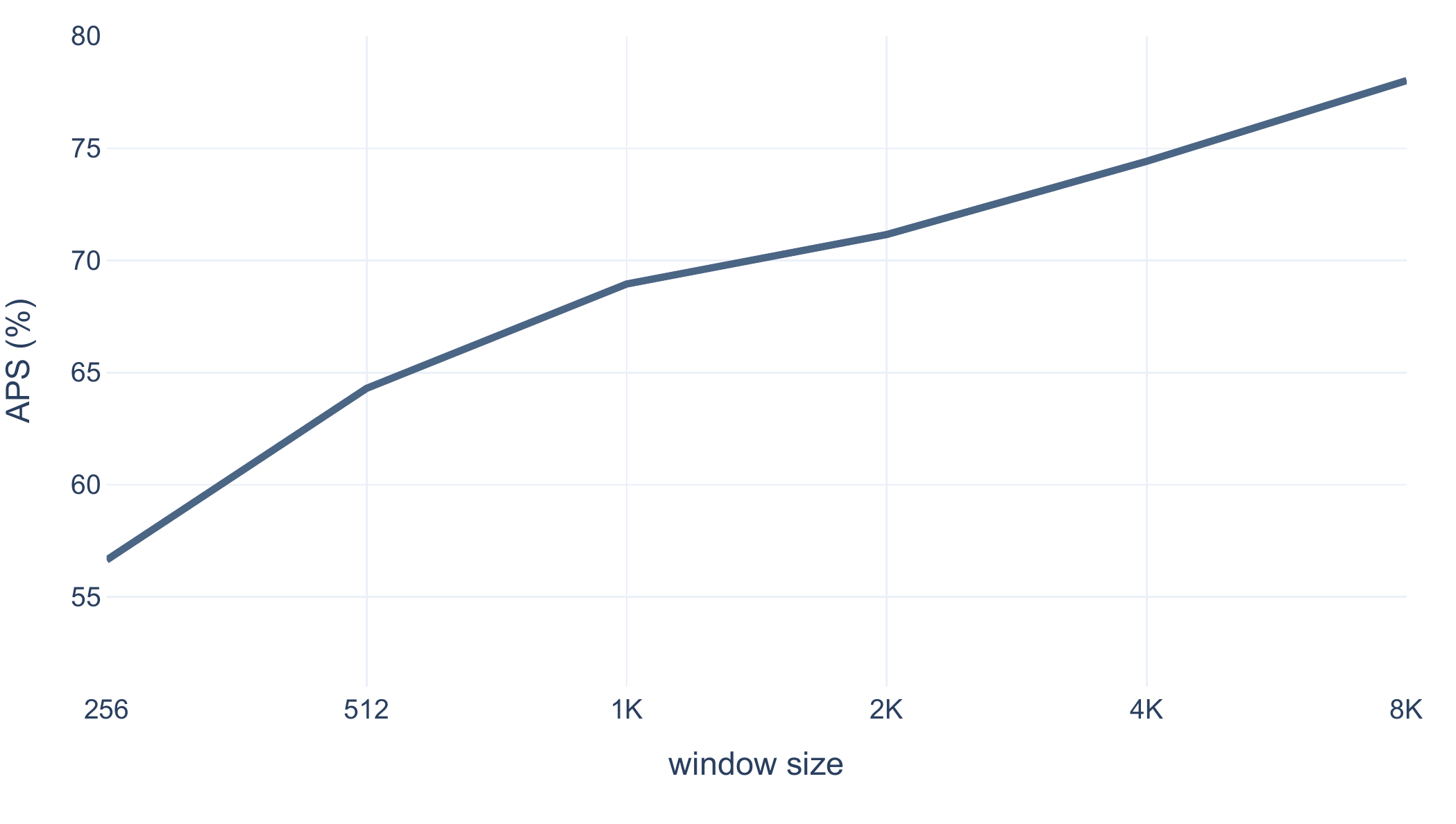}
    \caption{The MusicNet performance of our model on various window sizes. Using window sizes larger than 8K gives marginal accuracy improvement at a considerable increase in training time.}
    \label{fig:musicnet-lengths}
    \end{minipage}%
\end{figure*}

\begin{table}[ht!]
  \caption{Accuracy on LAMBADA word prediction task.}
  \label{lambada-comparison}
  \centering
  \begin{tabular}{lccc}
    \toprule
    Model                   & Parameters (M)     & Acc (\%)       \\
    \midrule
    Random word             & -                  & 1.6            \\
    Gated-Attention Reader  & unknown            & 49.0           \\
     SE                     & 33                 & 52.28          \\
    RSE (this work)         & 11                 & \textbf{54.34} \\
    Universal Transformer   & 152                & 56.0           \\
    Human performance       & -                  & 86.0           \\  
    GPT-3                   & 175000             & 86.4           \\
    \bottomrule
  \end{tabular}
\end{table}

In Fig~\ref{fig:lambada_time_comparison}, we compare the training and evaluation time of Residual Shuffle-Exchange (RSE), neural Shuffle-Exchange (SE) and Universal Transformer (UT) networks using configurations that reach their best test accuracy. We use the official Universal Transformer and neural Shuffle-Exchange implementations and measure the time for one training and evaluation step on a single sequence.
For the Universal Transformer, we use its base configuration with 152M learnable parameters. SE and RSE networks have 384 feature maps and 2 Bene\v{s} blocks, with total parameter count 33M and 11M, respectively.

We evaluate sequence lengths that fit in the 11GB of GPU memory. The Residual Shuffle-Exchange network works faster and can be evaluated on 4x longer sequences than Shuffle-Exchange network and 128x longer sequences than the Universal Transformer.

\subsection{MusicNet}

The music transcription dataset MusicNet \citep{thickstun2016learning} consists of 330 classical music recordings paired with the MIDI transcriptions of their notes. The total length of the recordings is 34 hours, and it features 11 different instruments. The task is to classify what notes are being played at each time step given the waveform. As multiple notes can be played at the same time, this is a multi-label classification task.

For performing classification, regularly spaced windows of a given length are extracted from the waveform. We predict all the notes that are played at the midpoint of the extracted window.

We use an RSE model with two Bene\v{s} blocks with 192 feature maps. We experimentally found this to be the best configuration. To increase the training speed, we prepend two strided convolutions in front of that, see analysis of other options below. To obtain the note predictions, we use the element in the middle of the sequence output by the RSE model, linearly transform it to the 128 values, one for each note pitch, and apply the sigmoid cross-entropy loss function to perform multi-label classification.

For evaluating the model, we use the average precision score (APS), which is the area under the precision-recall curve. This metric is well suited to prediction tasks with imbalanced classes and is suggested for the MusicNet dataset in the original paper \citep{thickstun2016learning}. We evaluate APS using the scikit-learn machine learning library \citep{pedregosa2011scikit}.

We train the model on different window sizes ranging from 128 to 8192 (see  Fig.~\ref{fig:musicnet-lengths}). We find that larger window sizes invariably lead to better accuracy. The best APS score of 78.02\% is obtained on length 8192.

The previous state-of-the-art was achieved by the Translation invariant net \citep{Thickstun2018InvariancesAD}. They used task-specific handcrafted filterbanks to achieve invariance in representation with respect to pitch shifts of the input audio. It incorporates the prior knowledge of invariances in the problem domain to train on an augmented version of the MusicNet, where data is pitch-shifted by an integral number of semitones.

See Table~\ref{musicnet-comparison} for a comparison with other works. A majority of the best results on the MusicNet have architectures that are specialized to use complex-valued data. This can give an advantage in sound processing tasks where data can be Fourier-transformed into a complex-valued representation before it is given as an input to the model. \citet{trabelsi2018deep} achieves 72.90\% APS with Deep Complex Network -- a complex-valued convolution architecture. Its real-valued counterpart Deep Real Network achieves a lower score of 69.80\% APS.
cgRNN is an RNN that uses a recurrent cell with complex-valued transitions \citep{wolter2018complex}. It achieves 53\% APS and uses only 2.36M parameters. \citet{yang2019complex} proposed Complex Transformer that achieves 74.22\% APS.

\begin{figure*}[!ht]
    \centering
    \begin{minipage}[t]{.49\textwidth}
    \includegraphics[width=\columnwidth]{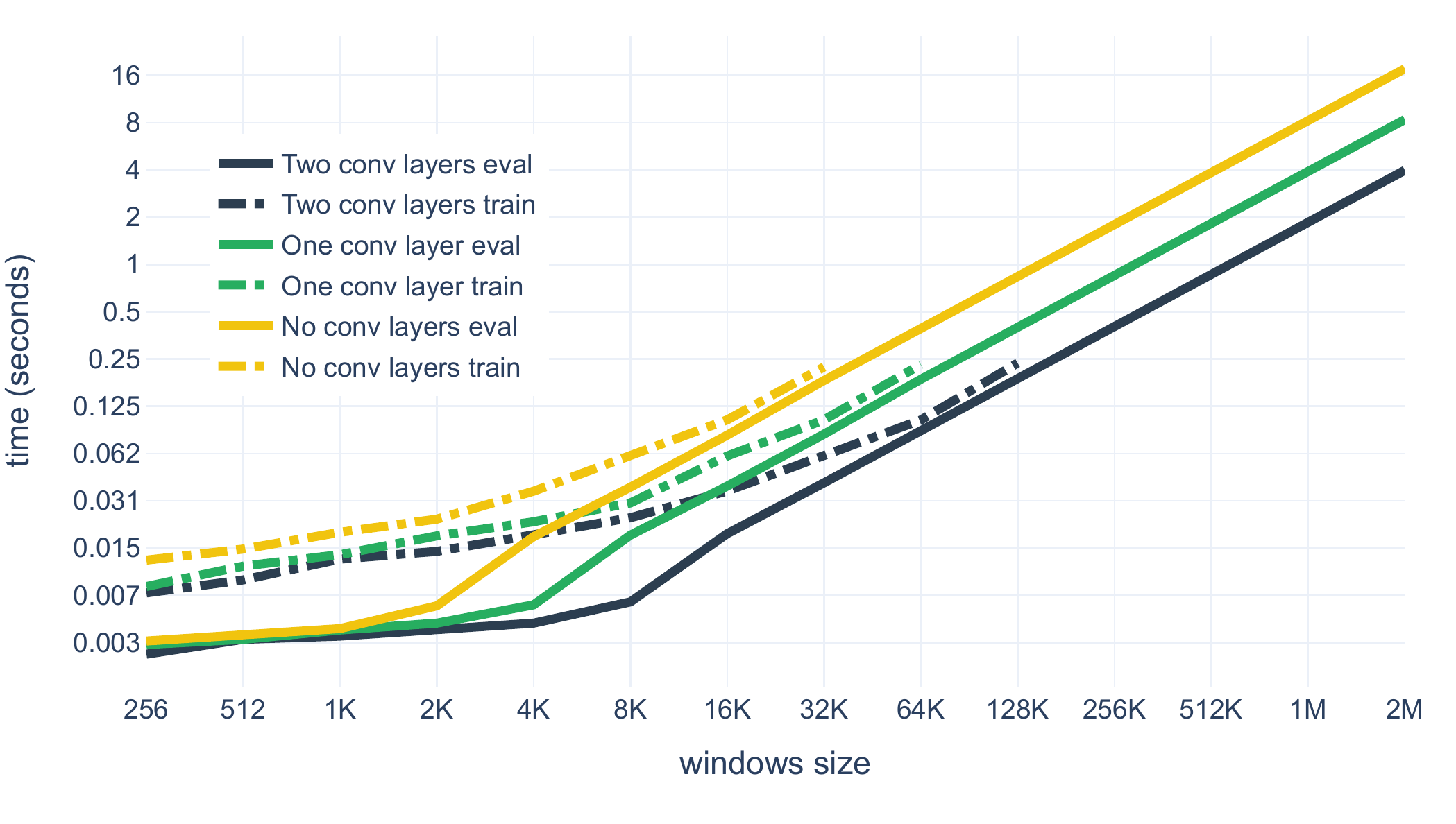}
    \caption{Training and evaluation speed (log-scale) depending on the prepended convolution count on the MusicNet task.}
    \label{fig:musicnet_conv_ablation}
    \end{minipage}%
    \hspace{.015\textwidth}
    \begin{minipage}[t]{.49\textwidth}
    \includegraphics[width=\columnwidth]{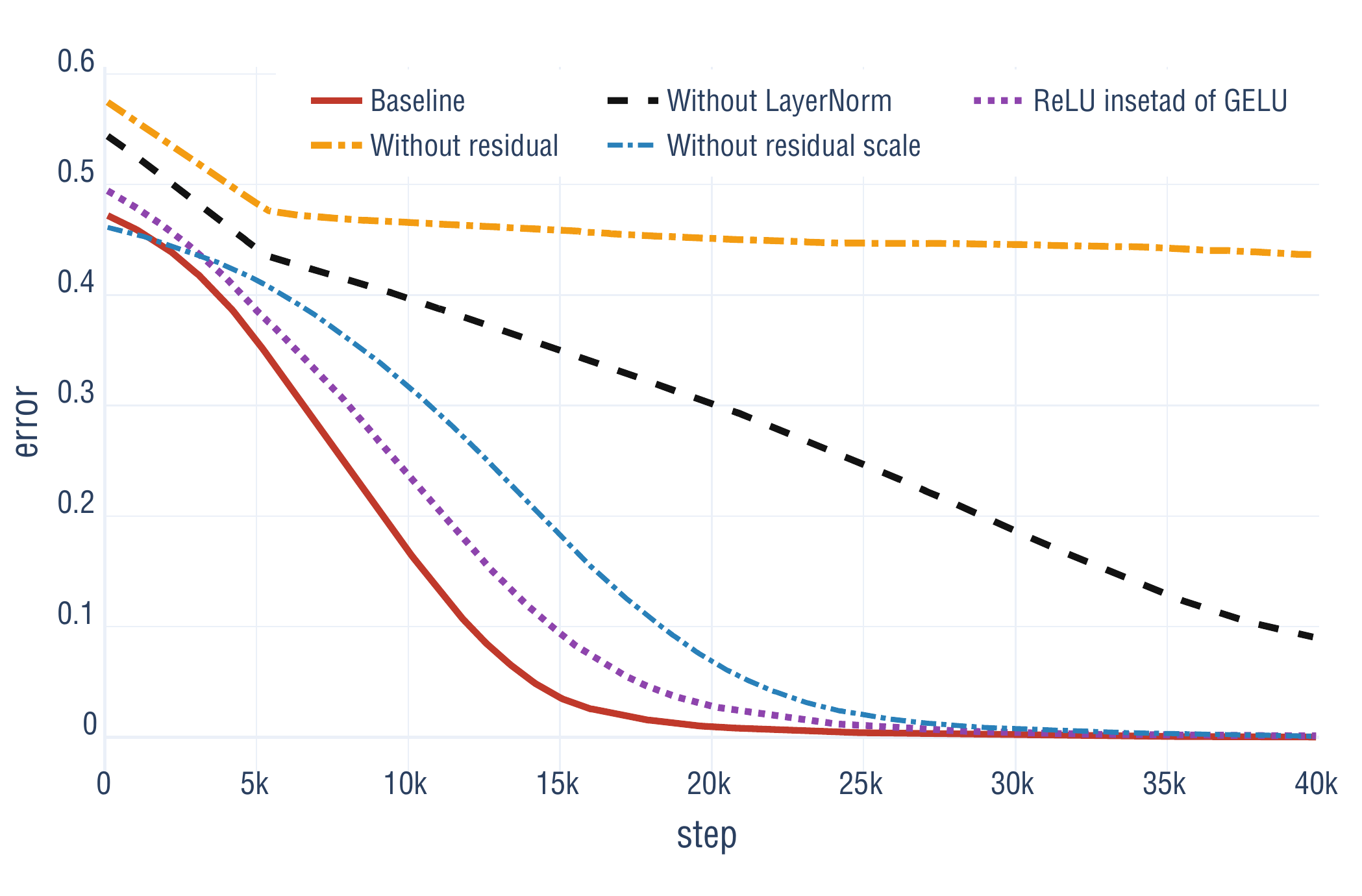}\hfill
    \caption{Ablation experiments. The plot shows test error on the multiplication task length 128 vs training step.}
    \label{fig:ablation}
    \end{minipage}%
\end{figure*}

\begin{table}[!htp]
  \caption{Performance on the MusicNet dataset.}
  \label{musicnet-comparison}
  \centering
  \begin{tabular}{lccc}
    \toprule
    Model               & Parameters (M)    & APS (\%) \\
    \midrule
    cgRNN                       & 2.36      & 53.00 \\
    Deep Real Network           & 10.0      & 69.80 \\
    Deep Complex Network        & 8.8       & 72.90 \\
    Complex Transformer         & 11.61     & 74.22 \\
    Translation-invariant net   & unknown   & 77.3 \\
    RSE (this work)             & 3.06      & \textbf{78.02} \\
    \bottomrule
  \end{tabular}
\end{table}

We have tested how prepending convolutions impact the speed and accuracy of the model. We find that one convolution gives the best results, although the differences are small. We chose to use two convolutions for a good balance between speed and accuracy.

We use the batch size of one example in this test to see the sequence length limit our model can be trained and tested on a single GPU. Fig.~\ref{fig:musicnet_conv_ablation} shows the training and evaluation speed of the model depending on the number of convolutions. Increasing the number of convolution layers improves training and evaluation speed, and the version with two convolutions can be trained on sequences up to length 128K. With 86 predictions per second as in the test set, our state-of-the-art model can perform inference 1.93 times faster than realtime. The training lines stop at the length at which the model does not fit in the GPU memory anymore. Testing lines reach a 2M technical limitation of our implementation.

In Appendix \ref{sec:appendix-musicnet-convolutions}, we provide accuracy measurements depending on the window size and the number of convolutions. We visualize the note predictions and describe the most common errors in Appendix \ref{sec:appendix-musicnet-visualization}. Additional training details are provided in Appendix \ref{sec:appendix-musicnet-training}.

\subsection{Ablation study}
\label{sec:ablation}

We have chosen the multiplication task as a showcase for the ablation study. It is a hard task which challenges every aspect of the model and performance differences are clearly observable. We use a model with 2 Bene\v{s} blocks, 192 feature maps and train it on length 128. We consider the following simplifications of the proposed architecture: 
\begin{itemize}
  \item removing LayerNorm (without LayerNorm)
  \item using ReLU instead of GELU
  \item removing the residual connection; the last equation of RSU becomes $[o_1, o_2] = c$ (without residual)
  \item setting the residual weight parameter $\sigma(S)$ to a constant 1 instead of a learnable parameter; the equation becomes $[o_1, o_2] = i +  h\odot c$ (without residual scale)
\end{itemize}
 We can see in Fig.~\ref{fig:ablation} that the proposed baseline performs the best. Versions without residual connection or normalization do not work well. Residual weight parameter and GELU non-linearity give a smaller contribution to the model's performance. 

We investigate the effect of the RSU hidden layer size on performance. Parameter count and speed of the model is directly proportional to the hidden layer size; therefore, we want to select the smallest size, which gives a good performance. By default, we use $2m$ feature maps where $m$ is the number of feature maps of the model. Versions with half as large or twice as large hidden layer size are explored. We discover that a larger hidden layer size leads to better performance. We consider the choice of $2m$ hidden layer size a good compromise between performance and parameter count. The experimental results for various hidden layer sizes are found in Appendix \ref{sec:appendix-ablation}.

We have performed ablation experiments also for LAMBADA and MusicNet tasks with similar conclusions, but the differences are much less pronounced.


\section{Conclusions}

We have proposed a simpler and faster version of the neural Shuffle-Exchange network. It has O($n$ log $n$) complexity and enables processing of sequences up to length 2 million where standard methods, like attention, fail. While keeping the overall successful connectivity structure of the Shuffle-Exchange network, we have shown that using residual connections instead of gated connections in its design, gives a significant boost to the training speed and achieved accuracy. Additionally, we have shown how to combine the model with strided convolutions that increase its speed and sequence length that can be processed.

The proposed model achieves state-of-the-art accuracy in recognizing musical notes directly from the waveform -- a task where the ability to process long sequences is crucial.
Notably, our architecture uses significantly fewer parameters than most of the previous best models for this task.

Our experiments confirm the Residual Shuffle-Exchange network as a useful building block for long sequence processing applications.


\section*{Acknowledgements}
We would like to thank the IMCS UL Scientific Cloud for the computing power and Leo Truk\v{s}\={a}ns for the technical support.
We sincerely thank all the reviewers for their comments and suggestions.
This research is funded by the Latvian Council of Science, project No. lzp-2018/1-0327.

\bibliography{bibliography}

\newpage
\onecolumn
\appendix
\section{Residual Switch Unit}
\label{sec:appendix-RSU}

\begin{figure}[ht]
    \centering
    \includegraphics[ width=0.6\columnwidth]{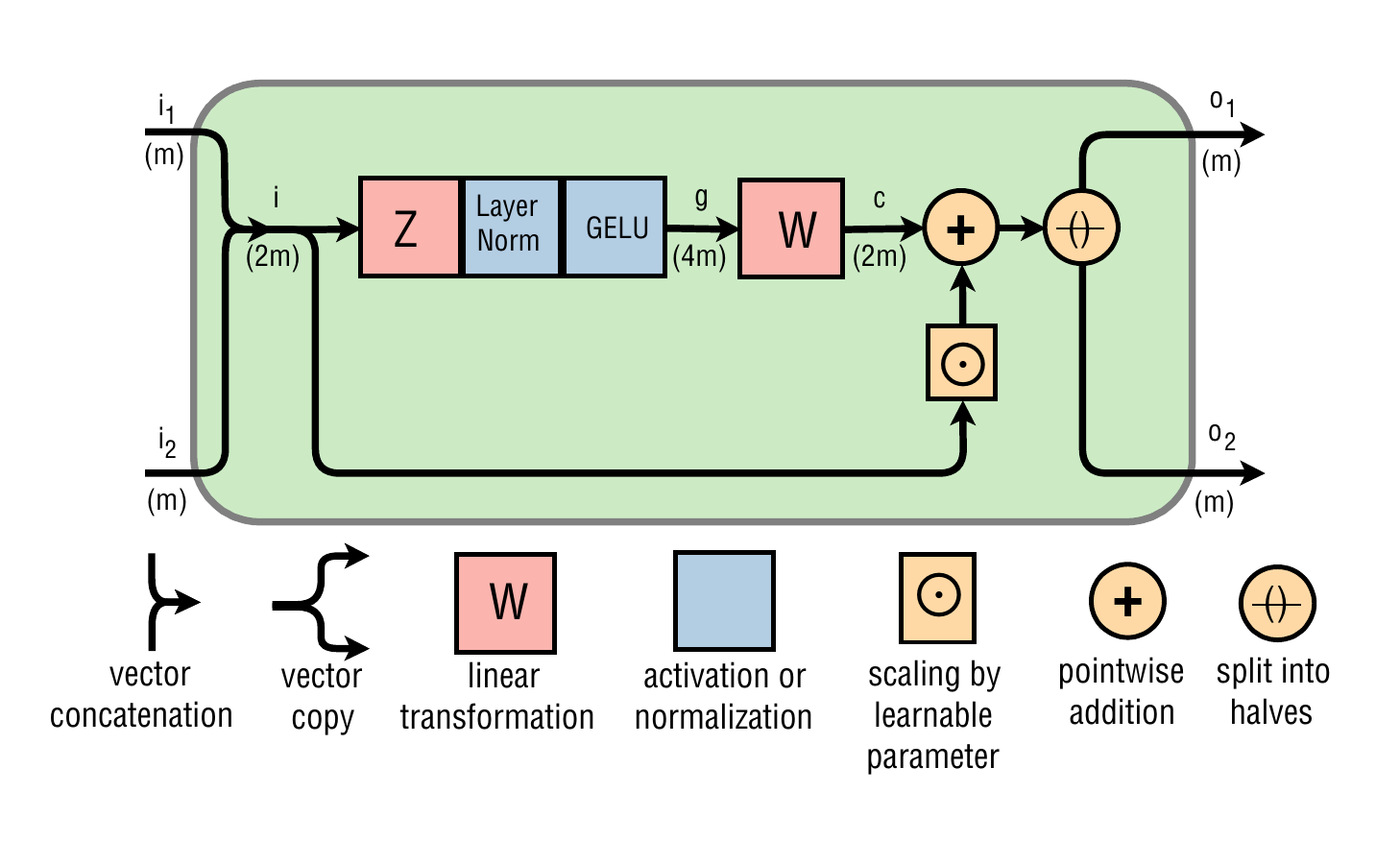}
    \caption{Residual Switch Unit. A number of feature maps ($m$) is shown in parentheses. Depicted here with the default of hidden layer being $2\times$ larger than the input ($4m$ being the size of the hidden layer and $2m$ the size of the input).}
    \label{fig:rsu-visualization}
\end{figure}

We visualize our Residual Switch Unit (see Fig.\ref{fig:rsu-visualization}) and place it in the context of the description found in section \ref{sec:model}. RSU consists of two linear transformations on the feature dimension. The first linear transformation is followed by Layer Normalization (LayerNorm) without output bias and gain \cite{xu2019understanding} and then by GELU. A second linear transformation is applied after GELU.
The RSU is defined as follows:

\centerline{$\begin{aligned}
i &= [i_1, i_2] \\
g &= \text{GELU}(\text{LayerNorm}(Z i))\\
c &= Wg + B\\
[o_1, o_2] &= \sigma(S)\odot i + h \odot c \\
\end{aligned}$}

In the above equations, $Z$, $W$ are weight matrices of size $2m \times 4m$ and $4m \times 2m$, respectively, $S$ is vector of size $2m$ and $B$ is a bias vector $-$ all of those are learnable parameters; $h$ is scalar value, $\odot$ denotes element-wise vector multiplication and $\sigma$ is the sigmoid function.

The rationale for the design of the Residual Switch Unit stems from there being two good design choices for deep networks (and our network is deep) - gated network (like LSTM) and residual network. Since the original unit was based on gates, we experimented with various choices based on a residual network. We found a design which is similar to the feed-forward block in the Transformer that works really well.

\newpage
\section{Algorithmic tasks}
\label{sec:appendix-algorithms}

\begin{figure}[ht]
    \centering
    \includegraphics[ width=0.6\columnwidth]{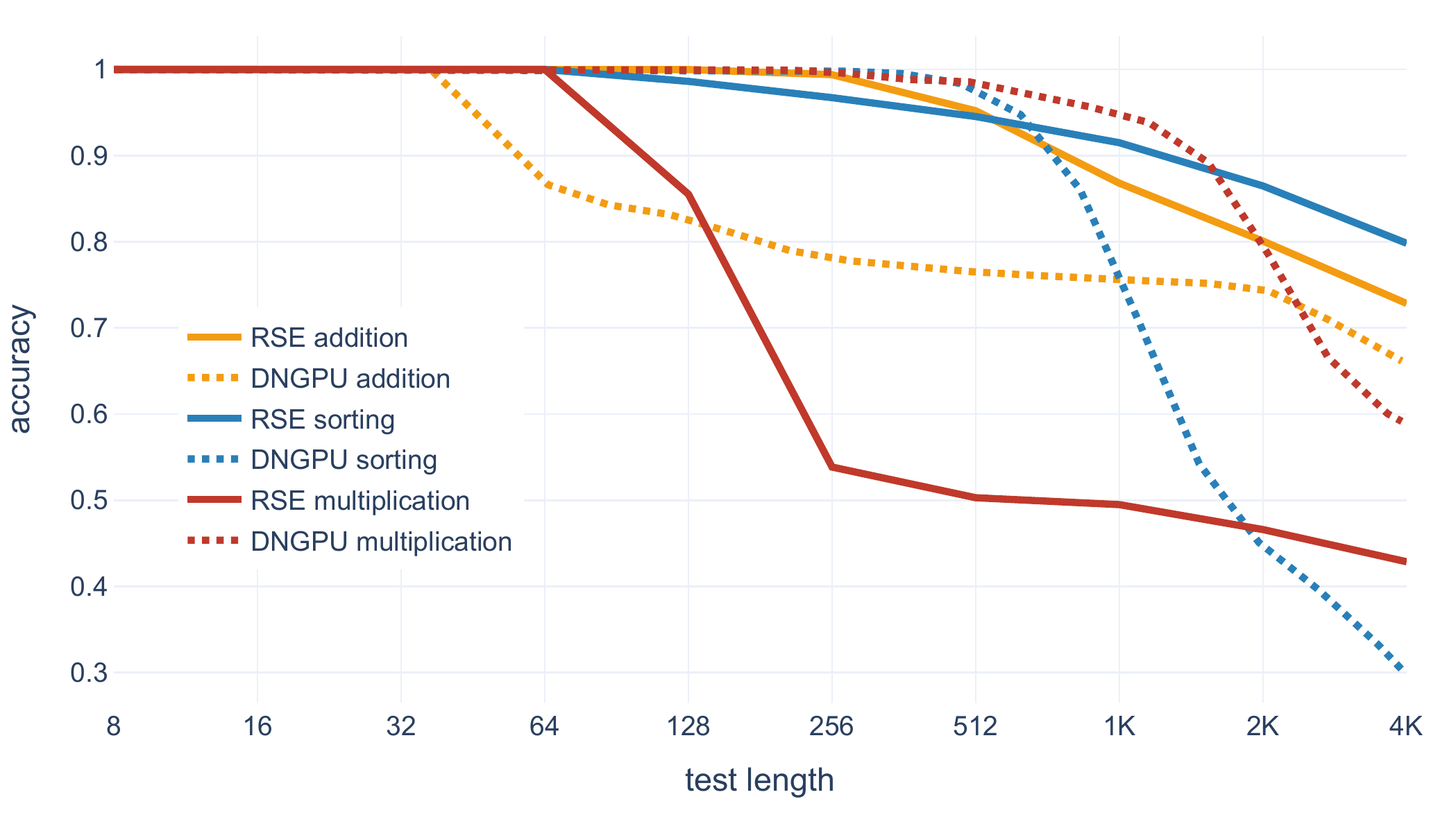}
    \caption{Test accuracy depending on the test length for the generalization of Residual Shuffle-Exchange (RSE) and Neural GPU with diagonal gates (DNGPU) to longer sequences on algorithmic tasks.}
    \label{fig:rse-vs-dngpu-gen}
\end{figure}

We compare the generalization of our proposed RSE with Neural GPU with diagonal gates (DNGPU). RSE generalizes better on sorting and addition tasks, but worse on the multiplication task (see Fig.\ref{fig:rse-vs-dngpu-gen}). Training the DNGPU is done as described by \citet{freivalds2017improving}.

\newpage
\section{Musicnet convolution count analysis}
\label{sec:appendix-musicnet-convolutions}

In the MusicNet dataset, each element of the input sequence is a single number. Residual Shuffle-Exchange network requires a large number of feature maps to work well, but encoding just one number into many feature maps is wasteful. For such tasks, we prepend the Residual Shuffle-Exchange network with several strided convolutions to increase the number of feature maps and reduce the sequence length.

\begin{figure}[ht]
    \centering
    \includegraphics[ width=0.6\columnwidth]{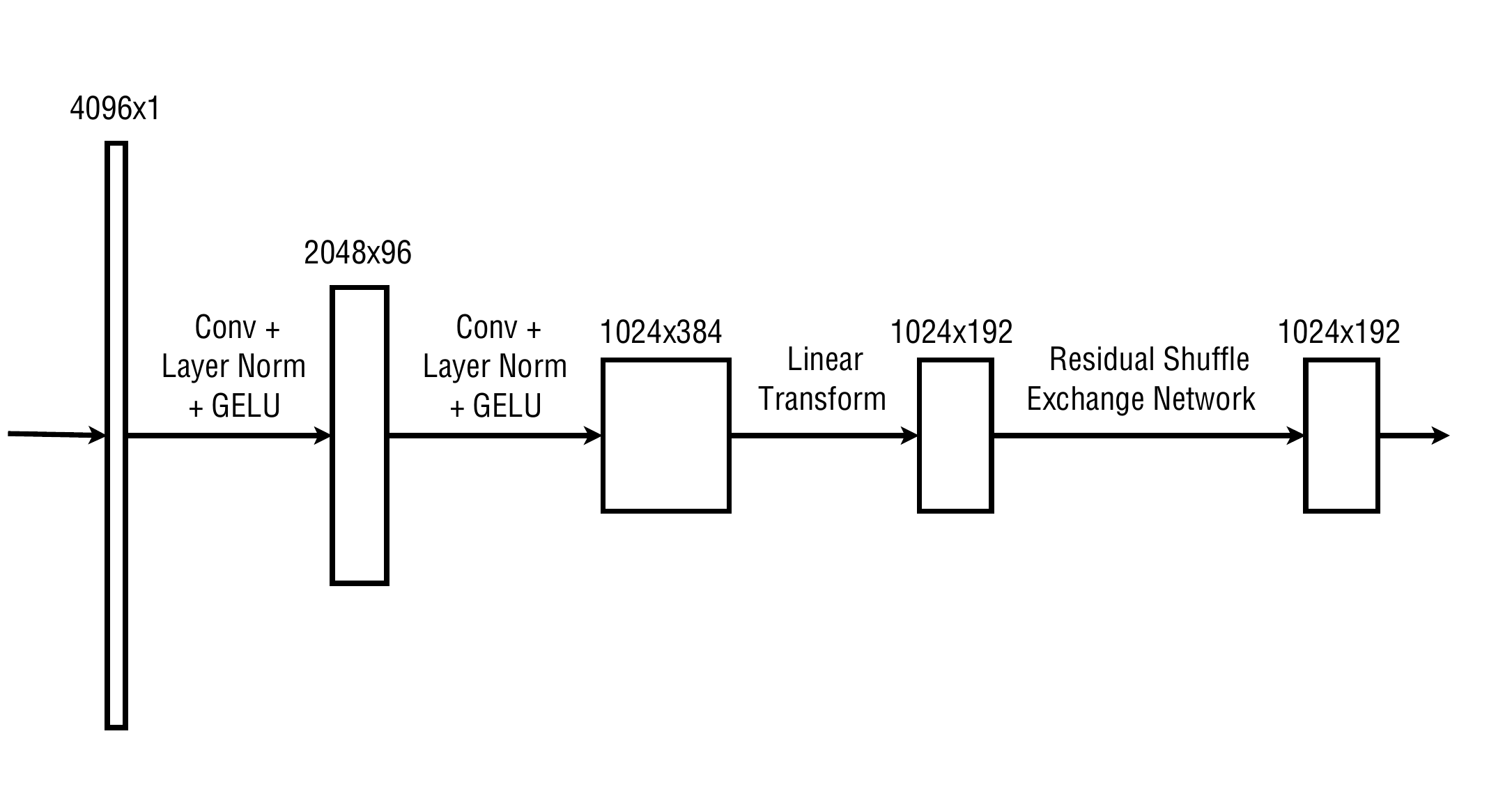}
    \caption{The architecture with two prepended convolutions employed for the MusicNet task.}
    \label{fig:musicnet-network}
\end{figure}

We use convolutions with stride 2 and apply LayerNorm and GELU after each convolution like in the RSU. Before the result is passed to the Residual Shuffle-Exchange network, a linear transformation is applied. In Fig. \ref{fig:musicnet-network} we illustrate an example model with two prepended convolution layers for an input sequence consisting of 4096 numbers.

We have tested how prepending convolutions impact the
speed and accuracy of the model. Table \ref{musicnet-convolutions} shows the obtained accuracy on window size 1024 for a different number of convolutions. We find that one convolution gives the best results, although the differences are small. We chose to use two convolutions for a good balance between speed and accuracy.

\begin{table}[!ht]
  \caption{Accuracy (APS) depending on the number of convolutional layers on length 1024.}
  \label{musicnet-convolutions}
  \centering
  \begin{tabular}{lccc}
    \toprule
    Convolutional layer count & APS (\%) \\
    \midrule
    0 & 69.29 \\
    1 & 69.57 \\
    2 & 68.95 \\
    3 & 67.39 \\
    \bottomrule
  \end{tabular}
\end{table}

\newpage
\section{MusicNet prediction visualization}
\label{sec:appendix-musicnet-visualization}

\begin{figure}[hbt!]
    \centering
    \includegraphics[width=0.6\columnwidth]{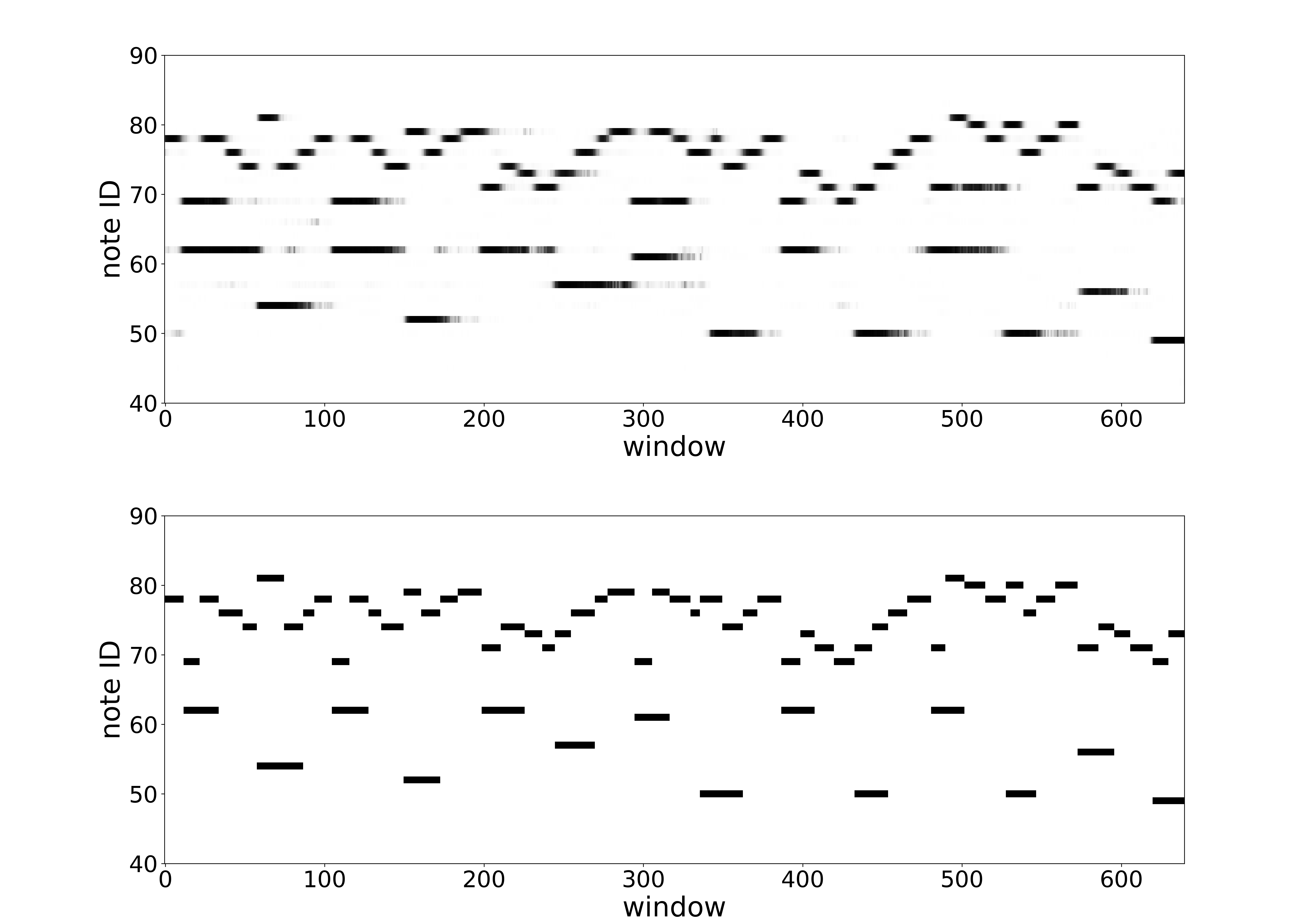}
    \caption{The top picture shows the predictions of our model on the test set. The predicted note shading represents the classification confidence. The bottom picture shows the corresponding labels. The vertical axis is the pitch of the note. The horizontal axis corresponds to the starting time of the audio segment that is given as the input to our model for predicting the notes played in the middle of the segment.}
    \label{fig:notes_visualisation}
\end{figure}

We visualize the note predictions that our model outputs with a window size 8192 (see Fig. \ref{fig:notes_visualisation}). This is the same model with which we achieved state-of-the-art accuracy. For creating input sequences for classification, regularly spaced windows of size 8192 are extracted from the waveform. Each window is shifted relative to the previous window by 128 elements of the sequence. The model predicts all the notes that are being played at the midpoint of the window.

The notes are generally well predicted but their start and end times are smoothed out. 
This affects low-pitched notes the most. The loss term described in Appendix \ref{sec:appendix-musicnet-training} was added to alleviate this by encouraging correct predictions at regularly spaced intervals of the input sequence. The added term decreased the number of errors at the edges of the notes and improved the overall performance.
In real applications, an appropriate threshold should be applied to get the duration of the notes.

\newpage
\section{MusicNet training details}
\label{sec:appendix-musicnet-training}

We train the model for 800K iterations which corresponds to approximately eight epochs. We found this value by examining the training dynamics on the validation set. The MusicNet dataset does not provide a separate validation set, so we split off six recordings from the training set and use them for validation. We use the same six recordings as \citet{trabelsi2018deep}. Using the same procedure as \citet{yang2019complex}, we downsample the waveform of the recordings in the dataset from 44.1 kHz to 11 kHz.

We add a term to the loss function, which predicts the notes played at regularly spaced intervals with stride 128 in the input sequence. This term is used only during training. The term is added because using only the middle element in the loss function seems to lead to lower accuracy in predicting the beginning and the end of the notes. The loss for these additional predictions is calculated in the same way as for the middle element.

We find that adding this term to the loss function improves the training speed and the final accuracy of the model. For example, without adding this loss term, we achieve 76.84\%, while with the added loss term, we achieve 78.02\%.

\begin{figure}[hbt!]
    \centering
    \includegraphics[width=0.6\columnwidth]{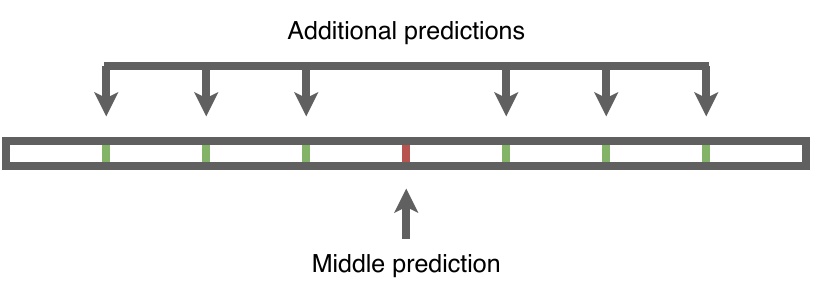}
    \caption{A window of size 1024 with additional predictions for the added loss term. The predictions are regularly spaced at sequence positions that are multiples of 128.}
    \label{fig:musicnet-loss}
\end{figure}

\newpage
\section{Additional ablation study}
\label{sec:appendix-ablation}

\begin{figure}[hbt!]
    \centering
    \includegraphics[width=0.6\columnwidth]{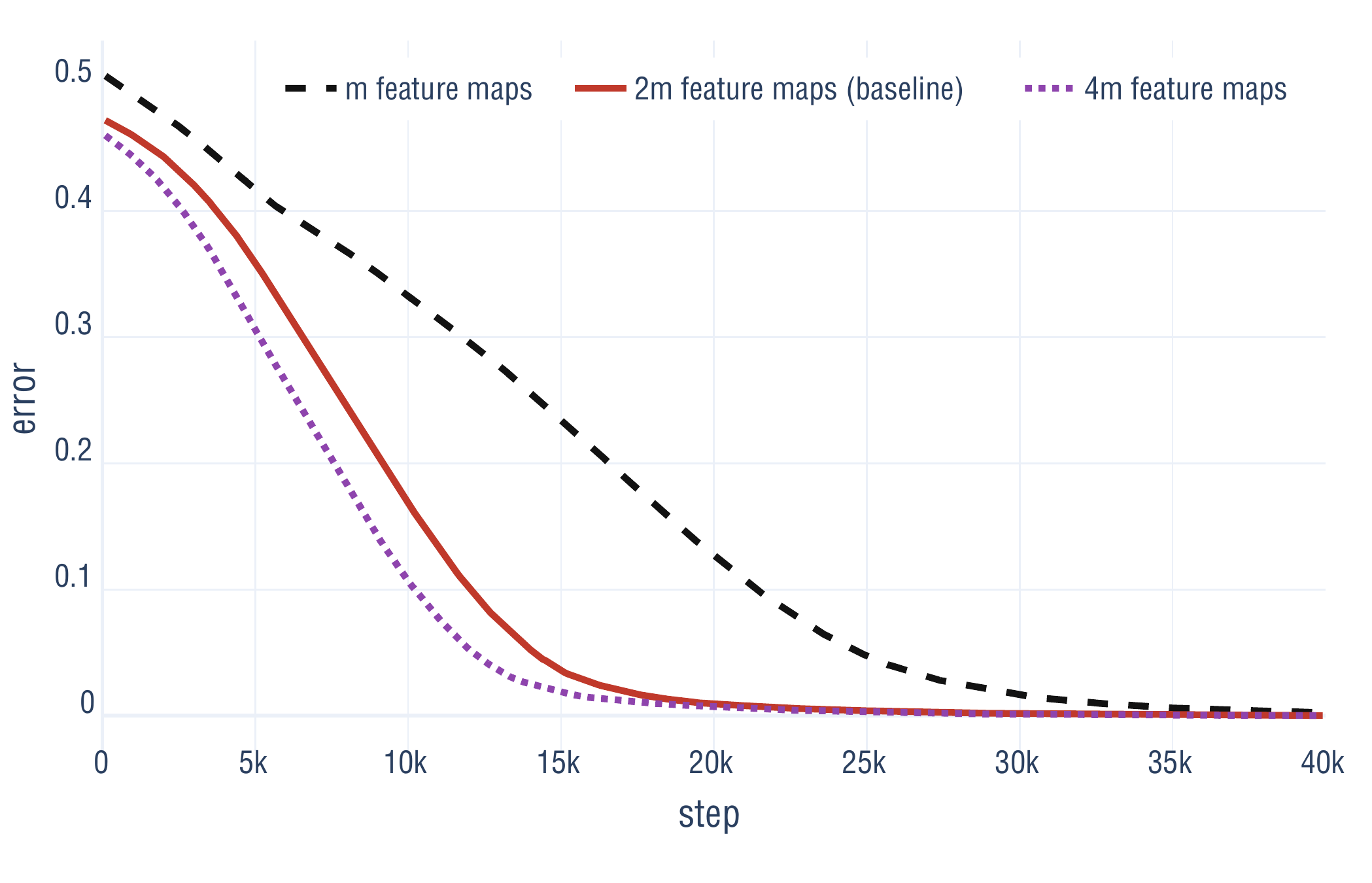}\hfill
    \caption{Test error depending on the hidden layer size ($m$ is the number of feature maps of the model).}
    \label{fig:hidden-size}
\end{figure}

We investigate the effect of the RSU hidden layer size on the performance. Parameter count and speed of the model is directly proportional to the hidden layer size; therefore, we want to select the smallest size, which gives a good performance. By default, we use $2m$ feature maps where $m$ is the number of feature maps of the model. Versions with half as large or twice as large hidden layers are explored. In Fig.\ref{fig:hidden-size} we can see that a larger hidden layer gives a better performance. We use a hidden layer of size $2m$ as a good compromise between performance and parameter count.

\end{document}